\documentclass{article}

\usepackage{microtype}
\usepackage{graphicx}
\usepackage{subfigure}
\usepackage{booktabs} %

\usepackage{hyperref}

\usepackage[final]{neurips_2025}

\usepackage{amsmath}
\usepackage{amssymb}
\usepackage{mathtools}
\usepackage{amsthm}

\usepackage[capitalize,noabbrev]{cleveref}

\theoremstyle{plain}

\theoremstyle{definition}

\theoremstyle{remark}

\usepackage[textsize=tiny]{todonotes}

\usepackage[utf8]{inputenc} %
\usepackage[T1]{fontenc}    %
\usepackage{url}            %
\usepackage{booktabs}       %
\usepackage{amsfonts}       %
\usepackage{nicefrac}       %
\usepackage{microtype}      %
\usepackage{xcolor}         %
\usepackage{graphicx}
\usepackage{multirow}
\usepackage{makecell}
\usepackage{graphicx}
\usepackage{subcaption}
\usepackage{enumitem}
\usepackage{titlesec}
\usepackage{bm}
\usepackage{xcolor}
\usepackage{tcolorbox}
\usepackage{mdframed}
\usepackage{algorithm}
\usepackage{algorithmic}

\titlespacing*{\paragraph}{0pt}{0.01ex plus 0.01ex minus .01ex}{0.7em}

\usepackage{amsmath}

\usepackage{tcolorbox}
\usepackage{fancyvrb}

\definecolor{mygrey}{HTML}{e8edf5}

\newtcolorbox{prompt}[1]{
    left=4mm,
    right=4mm,
    top=1.5mm,
    bottom=1.5mm,
    boxsep=0mm,
    rounded corners,
    title=#1,
    colback=mygrey, %
    colframe=black,    %
    fontupper=\footnotesize\linespread{0.85}\fontfamily{lmr}\selectfont,
    }

\newcommand{\ourmethod}{Diverse Preference Optimization}
\newcommand{\ourmethodsmall}{DivPO}
\newcommand{\piref}{\pi_\text{ref}}

\title{Diverse Preference Optimization}

\author{%
  Jack Lanchantin \\
  Meta \\
  \And
  Angelica Chen\\
  Meta, NYU
  \And
  Shehzaad Dhuliawala\\
  Meta, ETH Zürich
  \And
  Ping Yu \\
  Meta \\
  \AND
  Jason Weston \\
  Meta, NYU \\
  \And
  Sainbayar Sukhbaatar \\
  Meta \\
  \And
  Ilia Kulikov \\
  Meta \\
}

\begin{document}

\maketitle

\vspace{2pt}
\begin{abstract}
Post-training of language models, either through reinforcement learning, preference optimization or supervised finetuning, tends to sharpen the output probability distribution and reduce the diversity of generated responses.
This is particularly a problem for creative generative tasks where varied responses are desired. 
In this work we introduce {\em \ourmethod{} (\ourmethodsmall{})}, an optimization method which learns to generate much more diverse responses than standard pipelines, while maintaining the quality of the generations. In \ourmethodsmall{}, preference pairs are selected by first considering {\em a pool} of responses, and a measure of diversity among them, and selecting chosen examples as being more rare but high quality, while rejected examples are more common, but low quality. \ourmethodsmall{} results in generating
45.6\% more diverse persona attributes, and a 74.6\% increase in story diversity while maintaining similar win rates %
as standard baselines. On general instruction following, \ourmethodsmall{} results in a 46.2\% increase in diversity, and a 2.4\% winrate improvement compared to DPO.
\end{abstract}

\vspace{2pt}
\section{Introduction}
\vspace{2pt}

Large language models (LLMs) are proficient at producing high quality ``human-aligned'' outputs in response to a particular prompt \citep{touvron2023llama,team2023gemini,achiam2023gpt, jiang2024mixtral}. However, this alignment unfortunately results in a difficulty in producing a diverse set of outputs \citep{zhang2024forcing}. For example, repeatedly prompting a current state-of-the-art model to write a story with a particular title ends up producing stories with remarkably similar characters, events, and style. Aside from being an issue for individual user queries as just described, this also impacts the ability to generate high quality synthetic data -- 
which is becoming a vital component of model training via AI feedback,
where generated data from the model is fed back into the training loop, allowing for self-improvement \citep{wang2022self_i, bai2022constitutional, yuan2024self, singh2023beyond, chen2024self}. 

\begin{figure}[h]
    \centering
    \includegraphics[width=0.64\linewidth]{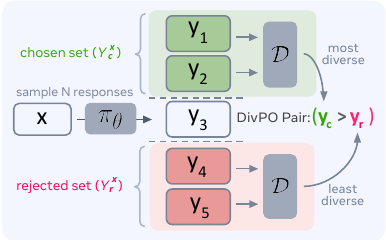}
    \vspace{6pt}
    \caption{\textbf{\ourmethod{}} (\ourmethodsmall{}). We consider a diversity criterion $\mathcal{D}$ for selecting chosen and rejected responses from a pool in preference optimization. Rather than taking the highest rewarded response as the chosen ($y_c$), we select the \textit{most diverse} response that meets a certain quality reward threshold. Similarly,  the \textit{least diverse} response that is below a threshold is selected as the rejected response ($y_r$). These are contrasted against each other to optimize both quality and diversity simultaneously.}
    \label{fig:divpo}
\end{figure}

The convergence of responses into a limited support distribution appears to stem from the model alignment phase, where the base language model is fine-tuned to align with human outputs and preferences \cite{kirk2023understanding,bronnec2024exploring, santurkar2023whose}. 
Model weights are tuned to optimize a reward (typically a proxy for human preferences). This results in the model placing a high probability on the highest rewarded response, and low on everything else. However, there may be other responses that have the same reward, but are ignored by the training loss. Ideally, we want responses with the same rewards to have the same probability of being generated. Further, when there is a small gap in reward between two responses, we also want their probabilities to be close. 

To address this limitation, we propose a novel training method called {\em \ourmethod{} (\ourmethodsmall{})} which aims to balance the distribution of quality responses given a prompt. 
The key intuition is that rather than contrasting the highest and lowest rewarded responses as typically done in preference optimization, we instead select the most diverse response that meets a reward (quality) threshold and contrast it with least diverse response that is below a reward threshold.
Our method is designed to not only achieve high rewards, reflecting alignment with human preferences, but also to maintain a high degree of diversity among the generated outputs. This dual objective is key for applications where both quality and variety are critical.

We experiment on two creative generation categories:  structured synthetic personas, and unstructured creative writing.
We first demonstrate how standard optimization approaches in state-of-the-art models increase reward but significantly reduce diversity. We then show how in contrast our method simultaneously increases the reward and the diversity compared to the baseline model. 
Our method is general and allows the use of any diversity criterion. In particular, we effectively generalize reward models to the case of assigning scores {\em given a pool of examples}, rather than being pointwise with independent scores.
We demonstrate that for any fixed quality target, \ourmethodsmall{} has higher diversity metrics than any baseline method. We find that on a general instruction following task, \ourmethodsmall{} not only leads to more diverse responses, but also higher quality compared to baselines.

\section{The Alignment Collapse Problem}
During the pre-training stage, language models are trained with a cross-entropy loss on diverse text corpora, resulting in a model that learns a distribution matching such data \citep{brown2020language, touvron2023llama}.
However, the ability of modern language models to respond to instructions in a manner preferred by humans is attributable to a post-training stage where they are ``aligned'' with human preferences over responses. Consequently, in this reinforcement learning from human feedback (RLHF) \citep{christiano2017deep} stage, the original learned distribution from the pre-training stage collapses. 
The reason behind this is the objective of reinforcement learning, which aims to optimize the cumulative future reward, $R$: $\mathcal{L} = -\sum_t r_t = -R,$
where $r_t$ can be seen as a reward for generating a token at time step $t$ in the case of text generation.
Putting all sequence-level probability mass on the highest reward point is an optimal solution to this loss. Even if multiple generations have the same reward, shifting all probability to only one of them is an optimal solution. Therefore, the objective of optimizing the cumulative reward causes collapse. 

To combat this, a regularizing KL term with respect to a reference model is often added to the RL loss:
\[
\mathcal{L} = -\sum_t r_t - \beta \text{KL}(\pi || \pi_\text{ref})
\]
This loss is used in both the PPO \citep{schulman2017proximalpolicyoptimizationalgorithms, ouyang2022training} and DPO \citep{rafailov2024direct} methods.
The optimal solution to this loss is
\[
\pi^* \propto \pi_\text{ref} \exp(R / \beta).
\]
This is more suitable because putting all probability on the highest reward is no longer optimal.
However, there can still be collapse depending on $\beta$.
Lower $\beta$ means higher reward generations can have even higher probabilities, thus leading to less diversity.
We cannot increase $\beta$ freely because it controls the KL term and higher $\beta$ will force the model to stay similar to the original model, which is less well aligned to human preferences. Yet despite the addition of this KL constraint, many works have showed that RLHF substantially decreases the diversity of language model outputs along several linguistic axes \citep{bai2022training,kirk2023understanding,guo2024benchmarkinglinguisticdiversitylarge,murthy2024fishfishseaalignment}, even when collaborating with humans \citep{padmakumar2024does}.

Furthermore, language models are typically evaluated on the ``quality'' of their responses, which ignores issues of generator collapse. Common evaluation metrics such as accuracy, pass@N \citep{kulal2019spoc,chen2021codex}, and win rate \citep{alpaca_eval,bai2022training} measure how frequently a language model's responses answer a query correctly, pass unit tests, or exhibit higher quality than another model's responses. These metrics can often be optimized even if the model always generates the same response for a given prompt, so long as the response is high quality. However, there are many tasks that benefit from \emph{both} high-quality and diverse responses. Tasks such as creative writing, idea generation, and biological sequence design, explicitly require and benefit from diverse generations \citep{marco2024smalllanguagemodelsoutperform,tachibana2025surprising,Madani2023diverse}. Further, recent work on inference scaling laws shows that LLM reasoning improves when searching a more diverse candidate space \citep{wang2022self, yao2024tree}. 
In addition, since LLMs are often used to generate synthetic training data, homogenous outputs can lead to significant downstream consequences, such as systemic bias \citep{yu2023diversitybias} and model collapse \citep{feng2024a,Shumailov2024}.

\section{Method}
\label{sec:method}
Given the issues with existing  preference optimization methods, we set forth two goals. First, we want high reward generations to be more likely than low reward generations, as is standard. Second, we want all high reward generations to have similar probabilities under the language model distribution.
Hence we introduce our method, called \ourmethod{} (\ourmethodsmall{}), which aims to optimize both of these goals simultaneously.

In  methods like DPO \cite{rafailov2024direct}, the goal is to optimize the reward margin, so the highest rewarded response is selected as the ``chosen'' response, and is typically contrasted with the least rewarded response, referred to as the ``rejected''\footnote{Other methods such as best-vs-random are also possible, but we use the common best-vs-worst approach.} \citep{yuan2024self, xu2023some, pace2024west}.

In \ourmethodsmall{}, we add a second constraint for selecting the chosen and rejected responses. Rather than selecting the highest rewarded response for the chosen, we want the \textit{most diverse} response that meets a certain reward threshold. Similarly, we want to reject the \textit{least diverse} response that is below a reward threshold. We loosely define a response as being more ``diverse'' if it differs substantially from other responses generated by the same model. We provide a more precise definition of diversity in the sections below.
Importantly, our method can accept any threshold and diversity criterion, which lets the user select their reward threshold tolerance and diversity measure that they want to optimize. 

Given this constraint, the steps for selecting a training pair for each training prompt are as follows. Given training prompt $x$, we first sample $N$ responses from initial model $\pi_\theta$ to create a pool of candidates, ${Y}^x=\{y_1,y_2,...,y_N\}$. We then score each response $y_i$ using a reward model, $s_i=\textsc{RM}(x, y_i)$. We then establish two buckets by thresholding the reward scores with a hyperparamter $\rho$: the chosen set ${Y}^x_c$ and rejected set ${Y}^x_r$, detailed in ~\autoref{sec:divpo_hyperparams}.

Next, we require diversity criterion (reward) $\mathcal{D}$ which in the general case takes as input a pool of responses ${Y}$ and outputs a diversity score for each: $d_i=  \mathcal{D}(y_i, {Y})$.
To determine the chosen response, we select the \textit{most diverse} response within the chosen set: $y_c = \textrm{argmax}_{y_i \in {Y}_c^x} \mathcal{D} (y_i, {Y}_c^x)$.
To determine the rejected response, we select the \textit{least diverse} response within the rejected set: %
$y_r = \textrm{argmin}_{y_i \in {Y}_r^x} \mathcal{D} (y_i, {Y}_r^x)$. 
We introduce several diversity criteria in ~\autoref{sec:divpo_hyperparams}.

We repeat this process for each training prompt to create a training set of preference pairs, as summarized in Alg.~\autoref{alg:example}. We use this set of diverse chosen and rejected responses to fit a Bradley-Terry model and update model $\pi_\theta$:
\begin{equation}\label{eq:optimum_model}
    \mathcal{L}_\text{\ourmethodsmall{}}(x; \pi_{\theta}, \piref) = -
    \log \sigma \bigg(\beta \log \frac{\pi_{\theta}(y_{c}\mid x)}{\piref(y_{c}\mid x)} 
    - \beta \log \frac{\pi_{\theta}(y_{r}\mid x)}{\piref(y_{r}\mid x)}\bigg).
\end{equation}
Here $\beta$ is used to control the deviation from a reference model $\piref$ (we use $\beta$=0.1 for all \ourmethodsmall{} experiments). In other words, we introduce a new preference pair selection method, and use the same optimizer as used in direct preference optimization \cite{rafailov2024direct}.

\subsection{\ourmethodsmall{} Configurations}
\label{sec:divpo_hyperparams}

\paragraph{Reward Threshold $\bm{\rho}$.}

To determine the chosen set $Y^x_c$ and rejected set $Y^x_r$, we introduce a hyperparameter $\rho$, which represents the percentage range from the lowest to the highest reward value.
All responses that have a reward value within $\rho$ percentage below the highest reward will be added to the chosen set. And all responses that have a reward value within $\rho$ percentage above the lowest reward will be added to the rejected set.  In other words, if $\rho = 0$, then the chosen and rejected responses will be identical to DPO, and if $\rho$=$0.5$, then all responses are considered.

\paragraph{Diversity Criterion $\bm{\mathcal{D}}$.} 

While our algorithm is general enough to allow for any diversity criterion $\mathcal{D}$, we use three different methods to determine the most and least diverse from a set of responses. 
\vspace{-5pt}
\begin{itemize}[leftmargin=0.5cm, itemsep=0.04em]
    \item \textbf{Model Probability.} If a response $y_i$ has a higher probability under the model, that means it is more likely to be generated again, hence less diverse.
    Thus we define $\mathcal{D}(y_i) = -\log \pi_{\theta}(y_i|x)$ so that less likely responses are considered more diverse. Note that this metric does not require a set of responses as input.
    \item \textbf{Word Frequency.} 
    Given a set of responses, we can measure how frequently a specific word occurs. A response with more frequent words is likely to be similar to other responses sharing the same words. Given this, we define $\mathcal{D}$ as inverse word frequency.
    \item \textbf{LLM-as-a-Diversity-Judge.} Finally, in general one could learn or predict diversity from a trained model, similar to reward modeling for quality. We prompt a language model to select the most and least diverse responses from the chosen and reject sets. See Appendix \autoref{fig:diversity_selection_prompt} for the prompt. 
\end{itemize}

\subsection{\ourmethodsmall{} Training}
\label{sec:divpo_training}
\ourmethodsmall{} can be used in both offline (off-policy) and online (on-policy) training. For online training the for loop in Alg.~\autoref{alg:example} is executed at every training step, but only over the current batch of prompts.
Compared to an offline setup, online training in other optimization approaches has shown performance improvements \citep{qi2024online,noukhovitch2024asynchronous} at the cost of computational efficiency.
In standard methods however online training is known to be more prone to collapse because as the model generation becomes less diverse, simultaneously so does the model's response training data. 
Our experiments include both offline and online setups which confirm the effectiveness of \ourmethodsmall{} regardless of the chosen training regime.

\begin{figure*}[ht]
    \centering
    \includegraphics[width=1.0\linewidth]{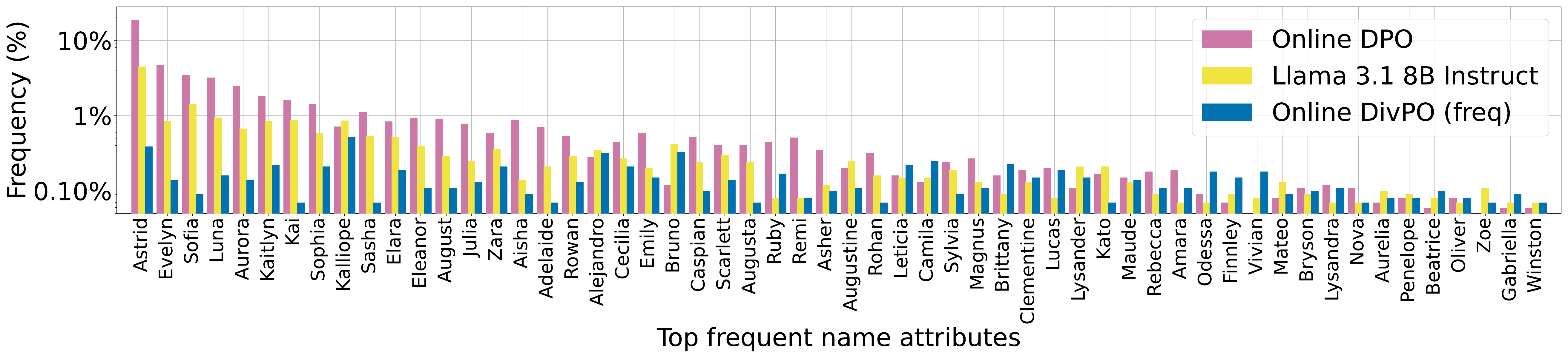}
    \caption{{\bf Persona Generation Statistics}.
    Llama 3.1-8B-Instruct and DPO tend to repeatedly generate a small subset of names, as shown by 
    frequency (\%) of the top most frequently generated. In contrast, \ourmethodsmall{} provides  
    a substantially more uniform distribution over the most frequent attributes, in addition to overall improved diversity metrics (see \autoref{tab:persona_performance_comparison}). }
    \label{fig:persona_hist_afterdivpo}
\end{figure*}

\section{Experiments}
\begin{table*}[h]
    \caption{{\bf Persona Generation Task Results.} 
    We report diversity and quality metrics comparing different training methods. Diversity for each attribute is defined as the number of unique attributes divided by the total number of generations satisfying the rule-based reward. Quality of generations is defined in two ways: the mean ArmoRM model score, and the percentage of valid JSON objects. The newly proposed \ourmethodsmall{} dramatically improves the diversity of attributes while maintaining the quality. Bold selections highlights best values independently for offline and online training rows.}
    \centering
        
    \resizebox{1.0\linewidth}{!}{%
    \begin{tabular}{lcccccc}
        \toprule
        & \multicolumn{4}{c}{\textbf{Diversity} $\uparrow$} & \multicolumn{2}{c}{\textbf{Quality} $\uparrow$} \\
    \cmidrule(lr){2-5} \cmidrule(lr){6-7}
    \textbf{Method} & \textbf{First name} & \textbf{City} & \textbf{Occupation} & \textbf{Avg} & \textbf{ArmoRM} & \textbf{Valid JSON \%} \\
        \midrule
        Llama 3.1-8B-Instruct & 30.45\% & 13.82\% & 27.93\% & 24.07\% & 0.141 & 45.51\% \\
        GPT-4o & 2.55\% & 0.74\% & 0.38\% & 1.22\% & 0.140 & 100.00\% \\
        \midrule
        SFT & 22.18\% & 9.14\% & 20.98\% & 17.43\% & \textbf{0.142} & \textbf{99.58\%} \\
        DPO & 22.95\% & 10.44\% & 27.92\% & 20.44\% & \textbf{0.142} & 99.25\% \\
        DivPO, $\mathcal{D}$=Freq & \textbf{49.68\%} & 27.87\% & 57.47\% & 45.01\% & 0.139 & 98.73\% \\
        DivPO, $\mathcal{D}$=Prob & 48.89\% & \textbf{29.86\%} & \textbf{58.44\%} & \textbf{45.73\%} & 0.139 & 97.32\% \\
        \midrule
        Online SFT & 24.92\% & 6.92\% & 19.46\% & 17.10\% & 0.139 & 99.54\% \\
        Online DPO & 11.61\% & 3.19\% & 10.82\% & 8.54\% & 0.139 & \textbf{99.99\%} \\
        Online DivPO, $\mathcal{D}$=Freq & 52.04\% & \textbf{45.35\%} & \textbf{65.03\%} & \textbf{54.14\%} & \textbf{0.141} & 99.80\% \\
        Online DivPO, $\mathcal{D}$=Prob & \textbf{53.85\%} & 29.21\% & 55.77\% & 46.28\% & 0.134 & 98.26\% \\
        \bottomrule
    \end{tabular}
    }
    
    \label{tab:persona_performance_comparison}
    \vspace{-5pt}
\end{table*}
We use the Llama 3.1-8B-Instruct model \cite{dubey2024llama} as the baseline model and as initialization checkpoint for training experiments. We utilize the fairseq2 library \citep{balioglu2023fairseq2} to implement \ourmethodsmall{} objective and execute supervised and preference fine-tuning recipes. We utilize the vLLM library \citep{kwon2023efficient} as the inference engine for data generation and evaluation. We use NVIDIA H100 GPUs for training and evaluation. Training hyperparameters can be found in \autoref{tab:training_details}.

\subsection{Persona Generation Task}
\label{subsec:personatask}

\paragraph{Task.} Our first task aims at generating a random character/person description by prompting the model to produce a JSON object containing various attributes. This is hence a constrained, structured output (with JSON specific keys).
Specifically, we choose 3 attributes: first name, city of birth, and occupation. \autoref{fig:persona_prompt} displays the prompt we use. 

\paragraph{Reward.} 
We define a {rule-based reward} based on the validity of the JSON output that can be used during training: a reward of 1 is assigned if the model's output contains a valid JSON object (and nothing else) featuring all the required attributes; otherwise a reward of 0 is assigned. 

\paragraph{Evaluation metrics.} 
We evaluate models by generating 10000 outputs conditioned on the prompt (ancestral sampling with temperature 1.0), and measure several metrics. %
We compute the {\em diversity} of each attribute as the number of unique attributes divided by the total number of generations satisfying the rule-based reward. The {\em quality} of the model's output is measured by the average score produced by the ArmoRM reward model \citep{wang2024interpretable} over the outputs satisfying the rule-based reward. ArmoRM takes as input the prompt and a model's response, and outputs a scalar value indicating response quality. Finally, we report the average rule-based JSON reward as another quality metric.

\paragraph{Diversity criterion.} 
When considering diversity of responses, we only consider the pool of valid JSON outputs.
Since all valid responses have the same reward of 1, we thus simply make the chosen and rejected pools equal to the all valid responses.
We experiment with using both the Probability and Word Frequency criteria for $\mathcal{D}$, as described in \autoref{sec:divpo_hyperparams}. For Frequency, we choose one attribute at random (out of name, city, occupation), and compute its frequency statistics over the valid responses. 
For Probability we use the length-normalized log probability to pick the chosen and rejected responses.
In both settings we also include extra preference pairs that feature random valid and invalid outputs (reward 1 and 0) as chosen and rejected targets in order to alleviate quality degradation during training. 

\paragraph{Preference training.}
In this task, we train models using both offline and online regimes. 
We do checkpoint selection for final evaluation using the rule-based reward value computed over a set of generations every 50 steps. We report checkpoint steps of each training run in \autoref{tab:model_selection_step_persona}.

\paragraph{Baselines.}
We consider two baseline methods: supervised finetuning (SFT) with NLL loss and preference finetuning using the DPO objective \citep{rafailov2024direct}. SFT uses Llama's generated outputs with reward 1 as targets, while DPO uses them as chosen and reward 0 as rejected.

\paragraph{Results.} 
While this task appears simple, standard state-of-the-art models 
such as Llama 3.1-8B-Instruct and GPT-4o
fail
to produce diverse personas (\autoref{tab:persona_performance_comparison}).
As shown in \autoref{fig:persona_hist_afterdivpo}, the Llama 3.1 model generates the name ``Astrid'' more than 10\% of the time, and in general is heavily skewed toward only a few names, as shown in Appendix \autoref{fig:persona_pretrained_llms_b}.
Adapting the sampling temperature does not alleviate this problem, 
as we see a sharp decline in quality, shown in Appendix \autoref{fig:persona_pretrained_llms_a}.

In contrast, we find  our newly proposed method substantially increases diversity in generated personas in both online and offline training regimes, while maintaining quality. 
As shown in \autoref{tab:persona_performance_comparison},
online \ourmethodsmall{} (using Word Frequency)  achieves an average (across attributes) improvement in diversity of {up to $30.07\%$} compared to the Instruct model and {up to $45.6\%$} compared to online DPO while achieving better quality in that case. The online \ourmethodsmall{} training regime shows up to a $9.14\%$ diversity improvement compared to its offline version. 
We find that both Diversity criteria (Frequency and Probability) achieve similarly strong performance, with slight wins for each in different settings.

In addition to generating more unique persona attributes, \ourmethodsmall{} improves the distribution over the most frequently generated attributes. \autoref{fig:persona_hist_afterdivpo} presents a histogram of the most frequently generated attributes compared to the Instruct model and DPO. 
The more uniform frequency distribution of \ourmethodsmall{} confirms that it helps to reshape the entire distribution rather than just lengthening its tail.

As reported previously in the literature, we confirm that DPO training is prone to diversity collapse. We also find that this effect is more severe in the online DPO setting compared to the offline one. However, online training shows greater stability during training as seen from both rule-based and ArmoRM reward measured during the training, with more detail shown in 
Appendix  \autoref{fig:persona-divpo-results}. We speculate that this might be related to the task's prompt distribution featuring a single prompt that makes learning saturate much faster during offline training compared to the online version.
Overall, DivPO works well in both offline and online cases, with stronger performance in the online setting.

\begin{figure*}[t]
    \centering
    \includegraphics[width=1.0\linewidth]{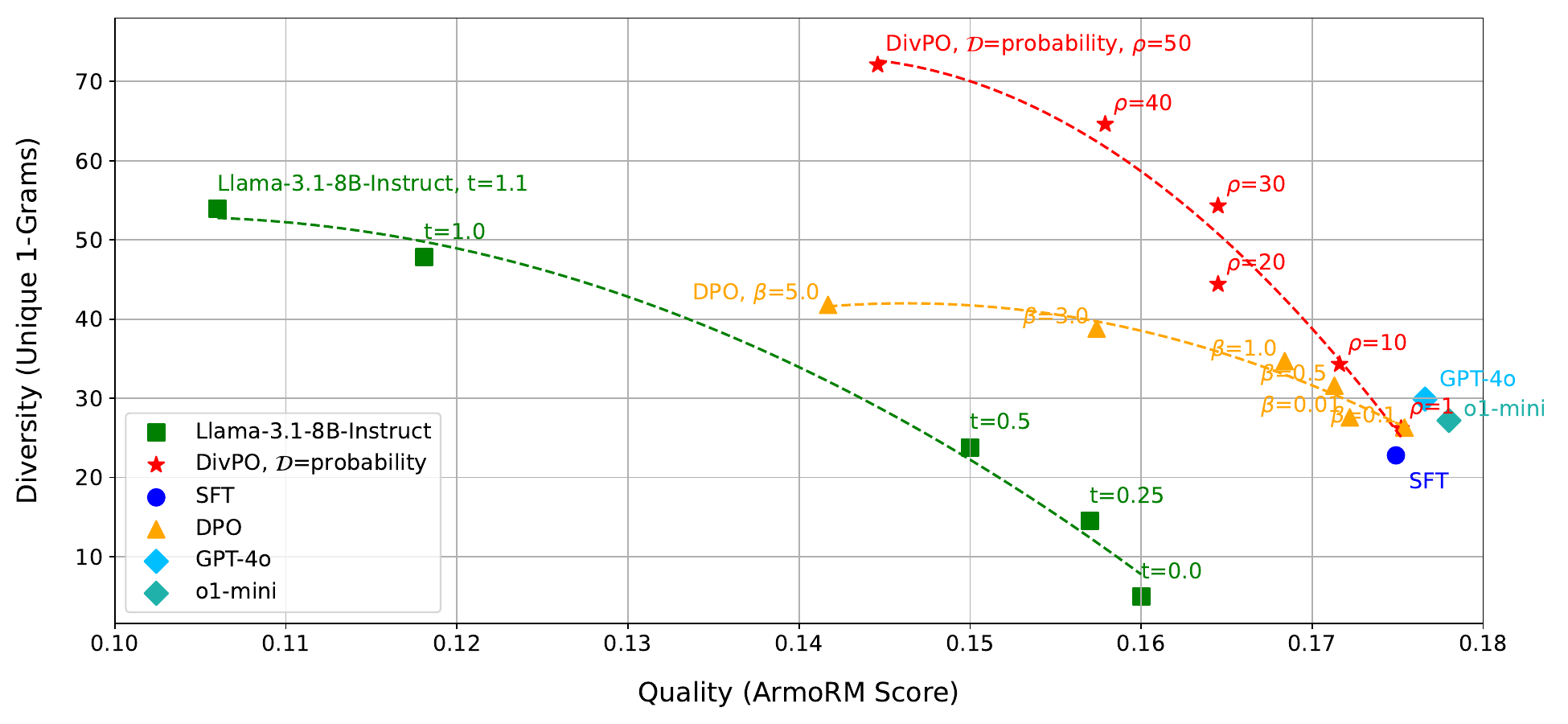}
\caption{{\bf Keyword Story Generation Results.} 
We show quality (ArmoRM scores) vs Diversity (Unique 1-Grams) for  $N$=16 responses per prompt. To tune diversity for baselines, we vary the baseline Lama 3.1-8B-Instruct temperature ($t$), and the DPO $\beta$ value. For our method, \ourmethodsmall{}, we vary the $\rho$ hyperparamter for choosing preference pairs. Unless otherwise noted, all methods use $t$=1. 
}
\label{fig:keyword_stories_results}
\vspace{-7pt}
\end{figure*}

\vspace{-5pt}
\subsection{Keyword Story Generation Task}
\label{sec:keyword_stories}

\paragraph{Task.} Many open ended prompts do not have a predefined set of valid outputs, and require a relatively unstructured creative output.
We therefore consider the setting of creative story writing. 
In this task we thus prompt the language model to write five keywords that could be used in a story with a particular title, as detailed in \autoref{fig:keyword_prompt}. Here, there is no specific set of valid outputs, but a requirement of exactly five words. 
We chose the five word story task for several reasons. First, its simple nature allows for systematic anlysis of our method in open ended generation. Second, restricting the output to five words allows us to easily measure diversity, which is non-trivial in open ended tasks \citep{tevet2020evaluating}.
Since there is no verifiable reward, we rely on a trained reward model to determine response quality. We create the dataset by first generating 6,000 synthetic story titles using Llama 3.1-405B-Instruct \citep{dubey2024llama}. Each of these are used in our prompt template (\autoref{fig:keyword_prompt}) to build our dataset prompts. We take 5k for training and 1k for testing. We then generate $N$=100 responses per training prompt with Llama3.1-8B-Instruct, which are used to create the training preference pairs (or SFT responses).

\paragraph{Reward.}
Following the Persona Generation task, we use the ArmoRM reward model to score responses. During evaluations, we manually set the reward value to 0 if the five word constraint is not met in the response since we observe that ArmoRM does not penalize missing the constraint.

\paragraph{Evaluation metrics.} For evaluation, we use three diversity metrics: compression ratio, unique 1-grams, and entropy. Details for how the metrics are computed are described in ~\autoref{sec:diversity_metrics}.
All diversity metrics are computed on the first five words.  For quality, we report  both the absolute ArmoRM score, as well as the ArmoRM win rate compared to the baseline Llama model.

\paragraph{Diversity criterion.} For \ourmethodsmall{}, we use the three diversity criteria outlined in ~\autoref{sec:divpo_hyperparams} (Probability-based, Frequency-based, or LLM-as-a-Diversity-Judge), and reward threshold values $\rho$ ranging from 0.01 to 0.5. For the Probability-based criterion, we use the log probability of all response tokens. For the frequency-based criterion, we first compute the frequency of each word in each set, $Y_c^x$ and $Y_r^x$, then compute the mean frequency of all words in each response $y_i$. 
We use the Llama 3.1-405B-Instruct model for the LLM-Judge case.

\paragraph{Preference training.} 
Each model is trained for 1k steps, with offline training (more efficient as we require LLM reward model calls). We use temperature$=$1 for all training and evaluation responses.

\paragraph{Baselines.} Similar to the Persona task, we consider two baseline training setups. Supervised finetuning (SFT) with NLL loss and preference finetuning using DPO. For each prompt, the top ArmoRM scored response is used for SFT, and best-vs-worst scored responses are used for DPO. We report varying both the quality/diversity tradeoff changing the $\beta$ parameter, and by varying decoding temperature.
We also evaluate GPT-4o \citep{hurst2024gpt} and o1-mini \citep{jaech2024openai}.

\paragraph{Results.} \autoref{fig:keyword_stories_results} shows the results for the \ourmethodsmall{} ($\mathcal{D}$=Prob) model compared to baselines, evaluating quality and diversity using mean ArmoRM reward scores and unique 1-grams, respectively. 
GPT-4o, o1-mini, and our 
SFT and DPO trained models 
all increase the quality (reward) compared to the baseline Llama 3.1-8B-Instruct model, but decrease the diversity  significantly. \ourmethodsmall{} similarly increases the quality compared to the base model, but can drastically increase the diversity compared to all baseline models. We show that we can control the amount of diversity or quality by changing the $\rho$ parameter in the \ourmethodsmall{} preference pair selection. The \ourmethodsmall{} ($\rho=0.3$) model achieves a 13.6\% increase in diversity and 39.6\% increase in quality compared to the Llama model. Compared to DPO, \ourmethodsmall{} with $\rho=0.3$ is 74.6\% more diverse, with only a drop of 6.4\% in quality. Importantly, \ourmethodsmall{} models always have \textit{equal or higher diversity at specific quality values} compared to the baselines.

We show the full results for all three diversity criteria $\mathcal{D}$=$\{$Probability, Frequency, LLM-Judge$\}$, and and all three diversity evaluation metrics (compression ratio, unique 1-grams, entropy) in Appendix \autoref{tab:fiveword}. \ourmethodsmall{} is more diverse and higher quality compared to the baseline Llama 3.1-8B-Instruct model for any diversity criterion. The $\mathcal{D}$=Prob model performs the best with a 35.1\% increase in unique 1-grams compared to the Llama 3.1-8B-Instruct model at a higher quality level. $\mathcal{D}$=Word Frequency and $\mathcal{D}$=LLM-Judge increase unique 1-grams by 15.7\% and 2\%, respectively, at a better quality level than the baseline model. Furthermore, for all criteria, we see a smooth tradeoff between diversity and quality by varying the $\rho$ parameter, demonstrating that whichever diversity criterion users prefer, they can change the diversity/quality threshold to achieve the right level of diversity.

Appendix \autoref{fig:keyword_example} shows an example test set prompt and the word count (overlap) statistics from the DPO and \ourmethodsmall{} ($\mathcal{D}$=Prob, $\rho=0.3$) models on $N$=16 generations for the story title \textit{``The Eyes of the World''}. The DPO model has a small number of unique words, and a highly skewed distribution within those. The words ``witness'', ``global'' and ``perspective'' dominate the 16 responses.
The \ourmethodsmall{} model has almost double the total amount of unique words, and a more uniform distribution
among them. 
Both models have a similar mean reward among the 16 generations, indicating that \ourmethodsmall{} learned a set of quality, yet diverse responses. We show the full set of responses in \autoref{fig:keyword_full_example}.

\paragraph{Full Story Generation Task.} Following the keyword stories experiments, we explore a full story generation task. Rather than the simpler task of generating a summary (5 keywords) in the previous section.
To do this, we utilize the generated keywords as seeds for crafting full stories. For each story title, we use $N$=16 keyword story generations derived from a given model.
For \ourmethodsmall{}, we use the $\mathcal{D}$=Prob models' seeds. 
Each of these keyword stories serves as a prompt for the Llama 3.1-8B-Instruct model, which then composes a full paragraph story, resulting in 16 stories per title. The template prompt to generate stories is provided in Appendix \autoref{fig:seeded_stories_prompt}. We evaluate the stories with the same diversity metrics used in the keyword story generation task. We use the ArmoRM model to evaluate story quality using the prompt without the keyword specification 
(i.e. only \texttt{Write a 1 paragraph story with the title \{title\}}). 

Results are shown in \autoref{tab:full_stories_results}.
We find a similar tradeoff between quality and diversity as in the previous task when varying the \ourmethodsmall{} $\rho$ parameter. Around $\rho=0.1$, we see similar quality between DPO and SFT, but higher diversity for \ourmethodsmall{}. For $\rho>0.1$, diversity improves further over the base model, with a slight drop in quality. We show an example of generated stories in Appendix \autoref{fig:seededstory_full_example}.

\subsection{Instruction Following Task}

\begin{table}[t]
\centering
\caption{Instruction Following Results. We show both diveristy and quality results for the AlpacaEval 2.0 benchmark, which measures general instruction following capabilities. \ourmethodsmall{} is both more diverse and higher quality than the baseline Llama 3.1-8b-instruct model and DPO finetuning.}
\vspace{4pt}
\resizebox{1.0\linewidth}{!}{%
\begin{tabular}{lccccccc}
    \toprule
    & \multicolumn{3}{c}{\textbf{Diversity} $\uparrow$} & \multicolumn{4}{c}{\textbf{Quality} $\uparrow$} \\
    \cmidrule(lr){2-4} \cmidrule(lr){5-8}
\raisebox{1ex}[0pt]{\textbf{Method}} & \textbf{\shortstack{Compr. \\ Ratio}} & \textbf{\shortstack{Unique \\ 1-Gram}} & \raisebox{1ex}[0pt]{\textbf{Entropy}} & \raisebox{1ex}[0pt]{\textbf{LC Winrate}} & \raisebox{1ex}[0pt]{\textbf{Winrate}} & \raisebox{1ex}[0pt]{\textbf{Std Err}} & \raisebox{1ex}[0pt]{\textbf{Length}} \\
    \midrule
    Llama 3.1-8b-instruct & 0.2497 & 1393.0 & 315.7 & 23.56 & 25.19 & 1.29 & 2643 \\
    DPO                     & 0.1894 & 994.7  & 127.0 & 41.30 & 38.83 & 1.42  & 1915 \\
    \midrule
    DivPO, $\mathcal{D}$=Prob, $\rho$=0.10 & 0.2207 & 1181.3 & 202.2 & 42.68 & 41.99 & 1.44  & 1978 \\
    DivPO, $\mathcal{D}$=Prob, $\rho$=0.20 & 0.2247 & 1177.5 & 189.0 & \textbf{44.07} & 39.66 & 1.44 & 1835 \\
    DivPO, $\mathcal{D}$=Prob, $\rho$=0.30 & 0.2464 & 1326.6 & 242.6 & 41.70 & 41.41 & 1.46 & 1985 \\
    DivPO, $\mathcal{D}$=Prob, $\rho$=0.40 & 0.2598 & 1454.6 & 284.1 & 42.29 & \textbf{42.50} & 1.46 & 2138 \\
    DivPO, $\mathcal{D}$=Prob, $\rho$=0.50 & \textbf{0.2949} & \textbf{1909.5} & \textbf{506.4} & 31.31 & 32.82 & 1.39 & 3878 \\
    \bottomrule
\end{tabular}
}
\label{tab:alpacaeval}
\end{table}

\paragraph{Task.} Lastly, we explore the effectiveness of using \ourmethodsmall{} on general instruction following tasks. We train on a random set of 10k single-turn Wildchat prompts \citep{zhao2024wildchat}, which is an open source dataset of 1 million real-world user-ChatGPT interactions.

\paragraph{Evaluation metrics.} We evaluate using the AlpacaEval 2.0 benchmark \citep{dubois2024length, alpaca_eval}, which uses GPT-4 as a judge to compute winrates against GPT4-turbo. This benchmark evaluates general instruction following capabilities on a diverse set of tasks. We sample 32 times for each prompt to compute diversity metrics (same as story generation). We then select 1 sample per prompt for AlpacaEval tests.

\paragraph{Diversity criterion.}  We use the Probability-based diversity criterion (since it is simple and shown effective in the toy task experiments), and reward threshold values $\rho$ ranging from 0.01 to 0.5.

\paragraph{Preference training.} Similar to story generation, we utilize the ArmoRM model to create preference pairs from $N=32$ samples per prompt. We train each model for a fixed 500 steps.

\paragraph{Baselines.} We compare to the seed Llama 3.1-8b-instruct model and vanilla DPO.

\paragraph{Results.} Results are shown in ~\autoref{tab:alpacaeval}. We find that \ourmethodsmall{} for all $\rho$ values between 0.10 and 0.40 are more diverse and higher quality than DPO.  Additionally, for $\rho=0.40$, the  \ourmethodsmall{} responses are significantly higher quality than the baseline Llama 3.1-8b-instruct model, with a similar or greater diversity (greater compression ratio and unique 1-gram diversity). 

The motivation of our method is to mitigate the diversity collapse that occurs when training with vanilla DPO. We can see that while DPO improves the quality compared to the baseline model, the diversity drops drastically. \ourmethodsmall{}, however does not have the same collapse, yet still improves the quality compared to both the baseline as well as the DPO model.
Our results indicate that training with a diversity constraint can not only mitigate collapse, but in fact lead to higher quality responses at test time because the model learns a wider variety of quality responses during training. These experiments demonstrate a promising result that the quality/diversity tradeoff does not necessarily have to hold when using \ourmethodsmall{}; we can simultaneously increase both quality and diversity.

For this task, we also include ablations of the N (samples per prompt) hyperparameter during training on the 10k Wildchat samples, shown in \autoref{tab:alpacaeval_ablations}. We ablate 3 different values of $N=\{16, 32, 64\}$ (samples per prompt) during training. For simplicity, we measure quality using ArmoRM on a set of 470 heldout prompts: 253 valid set examples from Humpback (Li et al. 2024) and 218 examples from the Evol-Test. (Xu et al. 2023). All evaluations are on 32 samples per prompt in the test set. We find that our method is robust to to all three $N$ sizes, where we show that DivPO (particularly for $\rho=\{0.1, 0.2\})$ outperforms DPO on all metrics for all $N$ values. We find that higher values of $\rho$ work better on smaller $N$.

\section{Related Work}
\paragraph{Preference Optimization and Collapse.}
Reinforcement Learning from Human Feedback (RLHF) is a crucial component of training LLMs that align with human preferences \citep{ouyang2022training}.
DPO \citep{rafailov2024direct} and other preference optimization methods \citep{xu2023some,meng2024simpo} have significantly simplified the RLHF process and yield similar improvements.
While these methods improve performance and generalization they can also negatively affect diversity and calibration \citep{achiam2023gpt,kirk2023understanding}.
In particular, RLHF methods optimize the final reward which does not take diversity into account, so it has become common practice to add a KL regularization term to maintain some of the model's original diversity \citep{ziegler2019fine,rafailov2024direct}.
\citet{wang2023beyond} find that RLHF under reverse KL divergence regularization can express a limited range of political views.
\citet{wang2023beyond} claim that the drop in diversity is because of the KL term. They test various f-divergences with DPO and show that the reverse KL gives the best accuracy but worst diversity, while forward KL gives best diversity but sacrifices accuracy. 

\paragraph{Mitigating Collapse During Training.} To promote diversity, \citet{li2015diversity} propose an alternative to SFT called Maximum Mutual Information, where they seek to maximize the pairwise mutual information between the source and the target. Similarly, \citet{li2024entropic} introduce a entropy-regularized alternative to cross-entropy in order to mitigate SFT collapse. 
\citet{zhang2024forcing} consider structured output tasks where they know an ideal distribution and introduce a supervised loss to match the model distribution with the ideal distribution. \citet{yu2024diversify} propose an iterative refinement method to resample instances from clusters during training. 
\citet{cideron2024diversity} include both CFG distillation and a diversity reward (the cosine similarity of embeddings) trained with RL, and apply it to music generation.
\citet{bradley2023quality} propose an evolutionary algorithm to discover unique (based on \citet{pugh2016quality}), yet quality responses and measure quality using a binary language model output. \citet{chung2023increasing} propose label replacement to correct misaligned labels in the training data. \citet{padmakumar2024beyond} propose a new method for collecting human preference data that encourages diversity. 
Lastly, several modifications to DPO were proposed to increase diversity.
\citet{wang2024preference} include multiple chosen and multiple rejected samples when creating preference pairs, but do not sample all responses from the same model or use an explicit reward model to allow for online self-improvement. %
\citet{park2024improving} augment DPO with an online diversity sampling term for maximizing protein structure stability and sequence diversity.
\citet{qin2025dive} propose a method for iterative self-improvement which re-uses previous model generations and improves the diversity of reasoning chains without compromising final answer accuracy.
To the best of our knowledge, we introduce the first method which directly modifies the preference pair selection process in order to simultaneously optimize quality and diversity.

\paragraph{Eliciting Heterogeneous Outputs During Inference.}

The simplest method for eliciting more or less heterogeneous outputs from a pretrained language model is to change the sampling distribution \citep{holtzman2019curious,fan2018hierarchical,nguyen2024turning, dhuliawala2024adaptive}. \citet{zhang2020trading} demonstrate the tradeoff between diversity and quality using sampling hyperparamters such as temperature, top-k, and top-p, and find that increasing all three perform similarly in order to increase diversity. However, sampling higher temperatures is known to lead to nonsensical output generations \citep{tevet2020evaluating}. \citet{eldan2023tinystories} enforce diversity by conditioning generations to incorporate three randomly chosen words into a story. However, this may not be applicable for all creative writing prompts.

\section{Conclusion}

In this paper, we introduced \ourmethod{} (\ourmethodsmall{}), a novel training method designed to address the lack of diversity in current state-of-the-art model responses, which is at least partially caused by existing preference optimization techniques. Our approach aims to add variety and balance the distribution of high quality responses by promoting diversity while maintaining alignment with high quality human preferences.

We demonstrated that standard optimization methods, while effective at increasing reward, often lead to a significant reduction in output diversity. In contrast, \ourmethodsmall{} successfully enhances both quality (reward) and diversity, as evidenced by multiple creative generation tasks, as well as general instruction following. 

Our method allows for users to define their own diversity criterion to be optimized, as well as a hyperparameter to control the diversity vs. quality tradeoff. \ourmethodsmall{} can be directly plugged into existing preference optimization frameworks, allowing users to easily achieve their target levels of output diversity. The main limitation of our method is that it requires a preference optimization loss, and does not work for other RLHF losses such as GRPO.
Future work could extend our method to further tasks, and integrate new methods of measuring diversity and rewards.

\bibliography{neurips_2024}
\bibliographystyle{icml2025}

\appendix
\clearpage

\begin{figure*}[h!]
    \centering
    \begin{prompt}{Persona generation} %
    \begin{Verbatim}[fontsize=\small]
Generate a random persona description with three characteristics.\n\n
Characteristics are:\n\n
- First Name\n
- The city of birth\n
- Current occupation\n\n 
Format the output strictly using JSON schema. Use `first_name` for First Name, 
`city` for the city of birth, `occupation` for current occupation as 
corresponding JSON keys. The ordering of characteristics should be arbitrary 
in your answer.
    \end{Verbatim}    
    \end{prompt}
    \vspace{-10pt}
    \caption{Instruction for the structured persona generation task.}
    \label{fig:persona_prompt}
\vspace{-5pt}
\end{figure*}

\begin{figure*}[ht]
    \centering
    \begin{prompt}{Keyword Story Generation} %
    \begin{Verbatim}[fontsize=\small]
List 5 words that could be used in a story titled "{title}". 
Do not write anything else but a list of 5 words without numbers.
    \end{Verbatim}    
    \end{prompt}
    \caption{Instruction template for the keyword story generation task.}
    \label{fig:keyword_prompt}
\end{figure*}

\begin{figure*}[t]
    \centering
    \begin{prompt}{Full Story Generation} %
    \begin{Verbatim}[fontsize=\small]
Write a 1 paragraph story with the title “{title}”. 
Your story must include the following elements: “{keywords}”.
    \end{Verbatim}    
    \end{prompt}
    \caption{Instruction template for the full story generation task.}
    \label{fig:seeded_stories_prompt}
\end{figure*}

\begin{figure*}[t]
    \centering
    \begin{prompt}{LLM Diversity Criterion Prompt} %
    \begin{Verbatim}[fontsize=\small]
You will be given a list of phrases, where each phrase is specified by an index 
number writtten as "[PHRASE_INDEX]". Select the index of the MOST UNIQUE phrase
compared to all other phrases. Consider unique words, word rarity, phrase
structure, etc. Do not explain your answer, just write the index.

Phrases:
{top_response_list}

Write your answer in the following format: "[PHRASE_INDEX]".
    \end{Verbatim}    
    \end{prompt}
    \caption{Prompt for selecting the most diverse keyword story out of a list of stories. A similar prompt is used for selecting the least diverse story, by replacing \texttt{MOST UNIQUE} with \texttt{LEAST UNIQUE}.}
    \label{fig:diversity_selection_prompt}
\end{figure*}

\section{\ourmethodsmall{} Details}

Algorithm ~\autoref{alg:example} outlines a detailed description of how to create \ourmethodsmall{} preference pairs for training.

\begin{algorithm}[tb]
   \caption{\ourmethodsmall{} Preference Pair Creation}
   \label{alg:example}
\begin{algorithmic}
   \STATE {\bfseries Require:} Training set $T$, diversity criterion $\mathcal{D}$, reward threshold $\rho$, base model $\pi_\theta$, reward model $RM$,
    
   \FOR{prompt $x$ in $T$:}
   \STATE 1. Sample $N$ responses: $\{y_1,y_2,...,y_N\} \sim \pi_\theta(x)$
   \STATE 2. Score each response: $s_i = \textsc{RM}(y_i,x)$
   \STATE 3. Determine chosen set $Y_c^x$ and rejected set $Y_r^x$ based on reward values and threshold $\rho$.
   \STATE 4. Use diversity criterion to find most diverse from the chosen set: $y_c = \textrm{argmax}_{y_i \in Y_c^x} \mathcal{D} (y_i, Y_c^x)$. Use diversity criterion to find least diverse from the rejected set: $y_r = \textrm{argmin}_{y_i \in Y_r^x} \mathcal{D} (y_i, Y_r^x)$.
    \STATE 5. Add $(x, y_c, y_r)$ to set of preference pairs
   \ENDFOR
\end{algorithmic}
\end{algorithm}

\section{Metrics}

\subsection{Diversity in story generation tasks}
\label{sec:diversity_metrics}

A diversity metric $M_d$, takes a set of model responses $Y$, and produces a scalar score of how diverse the set is \cite{kirk2023understanding}.

Given a model, $\pi$, we retrieve a set of $N$ outputs from the model for each prompt $x$ in the test prompt set $T$:
\[
 Y_{\pi}^{x} := \left\{y_{i} \sim \pi(y \mid x) \mid i = 1, \ldots, N\right\}
\]

For a diversity metric  $M_d$, we then evaluate the Prompt Level Diversity (the diversity of $Y_{\pi}^{x}$)
\begin{equation}
    \textrm{Prompt Level Diversity} := \frac{1}{|T|} \sum_{x \in T} M_d(Y_{\pi}^{x}) .
\end{equation}

For all metrics, we compute the prompt level diversity for the set of $N$ responses, and then average over all prompts in the test set.

\textbf{Compression Ratio.}
\cite{shaib2024standardizing} show that 
compression ratio (CR) is a surprisingly strong diversity metric that is more robust than other commonly used metrics. To compute compression rate for N responses, we first concatenate them into a single sequence, $[y_1,y_2,...y_N]$, then compress it using gzip. Finally, we compute the byte size ratio of the compressed sequence and original sequence:
\begin{equation}
    \textrm{Compression Ratio}= \frac{{\textsc{size}}(\textrm{gzip}([y_1,y_2,...y_N]))}{{\textsc{size}}([y_1,y_2,...y_N])}
\end{equation}

\textbf{Unique 1-Grams.} Unique 1-grams counts the number of distinct n-grams in the set of outputs. Since the tasks we consider in this paper are short sequences, we do not utilize $n > 1$ grams.

\textbf{Entropy.} Lastly, we consider the entropy of the responses, which can be estimated from the set of responses $\{y_1,y_2,...y_N\}$:
\begin{align}
\textrm{Entropy}_\pi(x) &= - \sum_y \pi(y|x) \log \pi(y|x) \\
&= \mathbb{E}_{y \sim \pi(y|x)} [-\log \pi(y|x)] \\
&\approx \frac{1}{N} \sum_{i=1}^N [-\log \pi(y_i|x)]
\end{align}

\subsection{Quality}

In the structured tasks, we have an explicitly defined measurement of response quality. If the response is a valid JSON output, then the quality/reward is 1, and if the response isn't a valid JSON, then the reward is 0. 

In the unstructured tasks, there is no easily verifiable measurement of quality. We therefore use the ArmoRM reward model \cite{wang2024interpretable}. This model takes as input a user response $x$ and a response $y$ and outputs a scalar value $r$ representing the quality of the response with respect to the prompt. In the keyword stories task, we manually set the reward to 0 if the word length constraint is not met.

\section{Additional Keyword Story Experiments}

\begin{table}[t!]
\caption{{\bf Full Story Generation Results.} We use the keywords generated by each model in the ``Method'' column as seeds for a Llama 3.1-8B-Instruct model to generate a full paragraph story. For \ourmethodsmall{} we use the $\mathcal{D}$=Prob model. Similar to the original keyword stories task, we observe a trend that \ourmethodsmall{} has higher diversity, while maintaining similar quality compared to the baseline Llama 3.1-8B-Instruct model.}
\centering
\resizebox{0.6\linewidth}{!}{%
\begin{tabular}{lcccc}
        \toprule
        & \multicolumn{3}{c}{\textbf{Diversity} $\uparrow$} & \multicolumn{1}{c}{\textbf{Quality} $\uparrow$} \\

\cmidrule(lr){2-4} \cmidrule(lr){5-5}
\raisebox{1ex}[0pt]{\textbf{Method}} & \textbf{\shortstack{Compr. \\ Ratio}} & \textbf{\shortstack{Unique \\ 1-Gram}} & \raisebox{1ex}[0pt]{\textbf{Entropy}} & \raisebox{1ex}[0pt]{\textbf{ArmoRM}} \\
        \midrule
        Llama3.1-8B-Inst. & 0.3663 & 765.6 & 193.0 & 0.1609 \\
        \midrule
        SFT & 0.3510 & 704.4 & 188.3 & 0.1633 \\
        DPO, $\beta$=0.1 & 0.3525 & 708.1 & 187.8 & 0.1638 \\
        \midrule
        DivPO, $\rho$=0.01 & 0.3530 & 709.4 & 187.5 & 0.1638 \\
        DivPO, $\rho$=0.1 & 0.3601 & 736.7 & 193.3 & 0.1625 \\
        DivPO, $\rho$=0.2 & 0.3698 & 783.9 & 206.4 & 0.1603 \\
        DivPO, $\rho$=0.3 & 0.3767 & 814.4 & 208.3 & 0.1600 \\
        DivPO, $\rho$=0.4 & 0.3855 &	844.2 &	216.2 &	0.1581 \\
        DivPO, $\rho$=0.5 & 0.3981 & 921.1 & 242.8 & 0.1528 \\
        \bottomrule
    \end{tabular}
}
\label{tab:full_stories_results}
\end{table}

\begin{figure}[ht]
    \centering
    \includegraphics[width=0.6\linewidth]{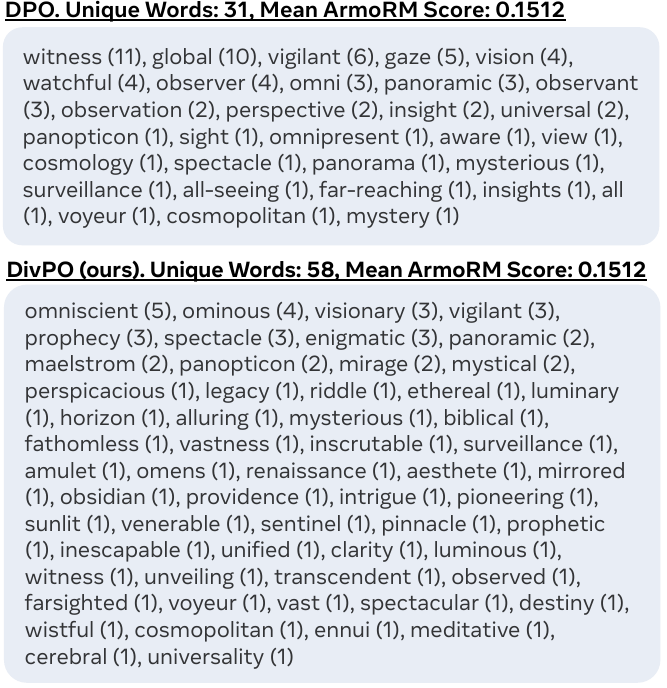}
    \caption{{\bf Keyword Stories Example Statistics}. We show word count statistics for $N$=16 generations of both the DPO and \ourmethodsmall{} ($\rho$=0.3, $\mathcal{D}$=Prob) model responses for the story title \textit{``The Eyes of the World''}. The DPO model has a small number of unique words, and a highly skewed distribution. \ourmethodsmall{}, has the same quality (ArmoRM score) as DPO with nearly double the amount of unique words, and a more uniform distribution among them.}
    \label{fig:keyword_example}
\end{figure}

\autoref{tab:fiveword} shows the full experimental results for the keywords stories tasks. We observe generalization of our \ourmethodsmall{} across different types of diversity criteria, $\mathcal{D}$.

\begin{table*}[h!]
\caption{Keyword Stories Results. 
We show the mean metric values for the $N$=16 responses per prompt. Winrates are compared to the baseline Llama 3.1-8B-Instruct model. We demonstrate three different diversity criteria: probability, word count, and LLM. \ourmethodsmall{} is both more diverse and achieves higher ArmoRM scores compared to the baseline model. \ourmethodsmall{} is significantly more diverse than DPO, while maintaining similar winrates.
}
\centering
\resizebox{1.0\textwidth}{!}{%
\begin{tabular}{lccccc}
        \toprule
        & \multicolumn{3}{c}{\textbf{Diversity} $\uparrow$} & \multicolumn{2}{c}{\textbf{Quality} $\uparrow$} \\

\cmidrule(lr){2-4} \cmidrule(lr){5-6}
\raisebox{1ex}[0pt]{\textbf{Method}} & \textbf{\shortstack{Compr. \\ Ratio}} & \textbf{\shortstack{Unique \\ 1-Gram}} & \raisebox{1ex}[0pt]{\textbf{Entropy}} & \raisebox{1ex}[0pt]{\textbf{ArmoRM}} & 
\textbf{\shortstack{ArmoRM \\ Winrate}} \\
\midrule
Llama 3.1-8B-Instruct & 0.498  & 47.8 & 20.3 & 0.1178 & - \\
GPT-4o & 0.326	& 29.9	& - & 0.1766 & 0.8715 \\
o1-mini & 0.361 & 	27.2	 & -	& 0.1780 & 0.8877 \\
\midrule
SFT & 0.316 & 22.8 & 5.8 & 0.1752 & 0.8290 \\
DPO & 0.396	& 31.1	& 14.0	& 0.1759 & 0.8351 \\
\midrule
DivPO, $\rho$=0.01, $\mathcal{D}$=Prob & 0.362 & 26.2 & 11.1 & 0.1752 & 0.8787 \\
DivPO, $\rho$=0.1, $\mathcal{D}$=Prob & 0.412 & 34.3 & 15.0 & 0.1716 & 0.8410 \\
DivPO, $\rho$=0.2, $\mathcal{D}$=Prob & 0.461 & 44.4 & 19.9 & 0.1645 & 0.7586 \\
DivPO, $\rho$=0.3, $\mathcal{D}$=Prob & 0.518 & 54.3 & 24.7 & 0.1645 & 0.6774 \\
DivPO, $\rho$=0.4, $\mathcal{D}$=Prob & 0.562 & 64.6 & 31.2 & 0.1579 & 0.5367 \\
DivPO, $\rho$=0.5, $\mathcal{D}$=Prob & 0.597 & 72.1 & 36.8 & 0.1446 & 0.3973 \\
\midrule
DivPO, $\rho$=0.01, $\mathcal{D}$=Freq & 0.375 & 28.0 & 11.9 & 0.1711 & 0.8560 \\
DivPO, $\rho$=0.1, $\mathcal{D}$=Freq & 0.413 & 34.9 & 14.6 & 0.1691 & 0.8471 \\
DivPO, $\rho$=0.2, $\mathcal{D}$=Freq & 0.458 & 43.0 & 18.2 & 0.1675 & 0.7909 \\
DivPO, $\rho$=0.3, $\mathcal{D}$=Freq & 0.471 & 43.7 & 17.8 & 0.1652 & 0.7230 \\
DivPO, $\rho$=0.4, $\mathcal{D}$=Freq & 0.476 & 45.6 & 18.3 & 0.1654 & 0.6883 \\
DivPO, $\rho$=0.5, $\mathcal{D}$=Freq & 0.540 & 55.3 & 22.5 & 0.1574 & 0.5502 \\
\midrule
DivPO, $\rho$=0.01, $\mathcal{D}$=LLM & 0.357 & 25.9 & 10.9 & 0.1762 & 0.8840 \\
DivPO, $\rho$=0.1, $\mathcal{D}$=LLM & 0.393 & 31.4 & 13.4 & 0.1575 & 0.7908 \\
DivPO, $\rho$=0.2, $\mathcal{D}$=LLM & 0.428 & 36.7 & 15.4 & 0.1673 & 0.7857 \\
DivPO, $\rho$=0.3, $\mathcal{D}$=LLM & 0.456 & 42.3 & 17.7 & 0.1662 & 0.7167 \\
DivPO, $\rho$=0.4, $\mathcal{D}$=LLM & 0.496 & 48.8 & 20.6 & 0.1622 & 0.5997 \\
DivPO, $\rho$=0.5, $\mathcal{D}$=LLM & 0.543 & 56.9 & 24.4 & 0.1545 & 0.4819 \\

\bottomrule
\end{tabular}
}

\label{tab:fiveword}
\end{table*}

\begin{figure*}
    \centering
    \includegraphics[width=0.6\linewidth]{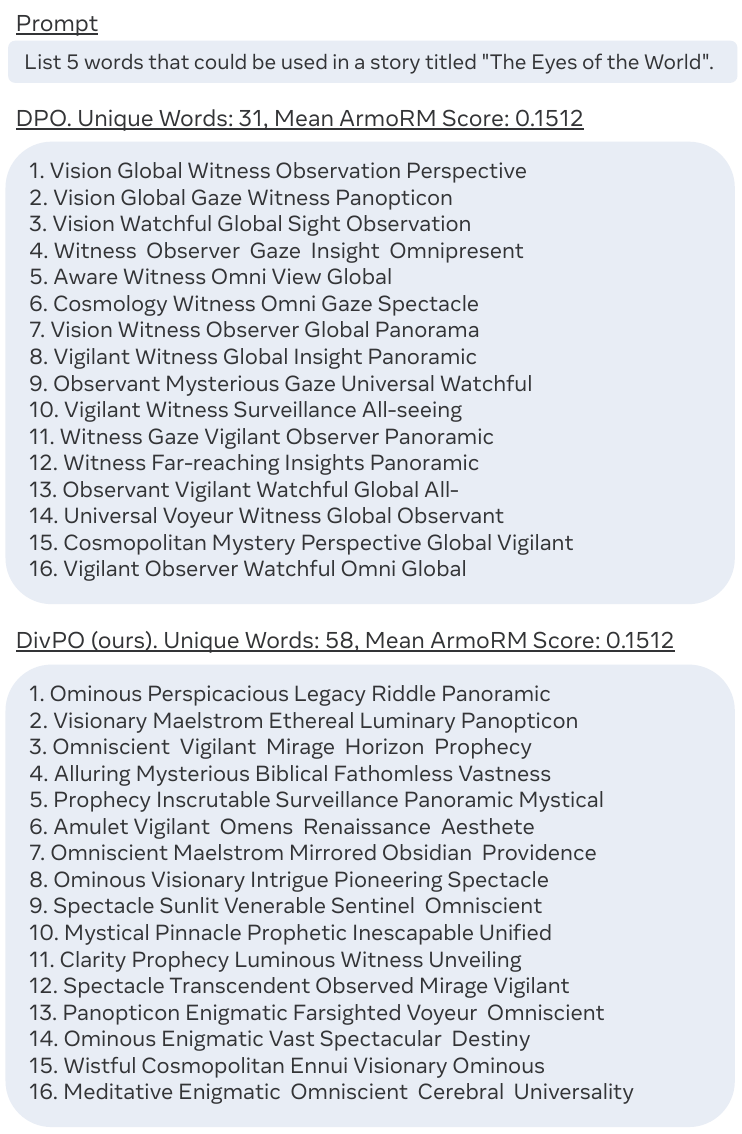}
    \caption{{\bf Keyword Story Example}. Here we show the full set of $N$=16 generations for one of the test prompts on the keyword stories task. \ourmethodsmall{} is more diverse, with similar quality compared to DPO.}
    \label{fig:keyword_full_example}
\end{figure*}

\begin{figure*}
    \centering
    \includegraphics[width=1\linewidth]{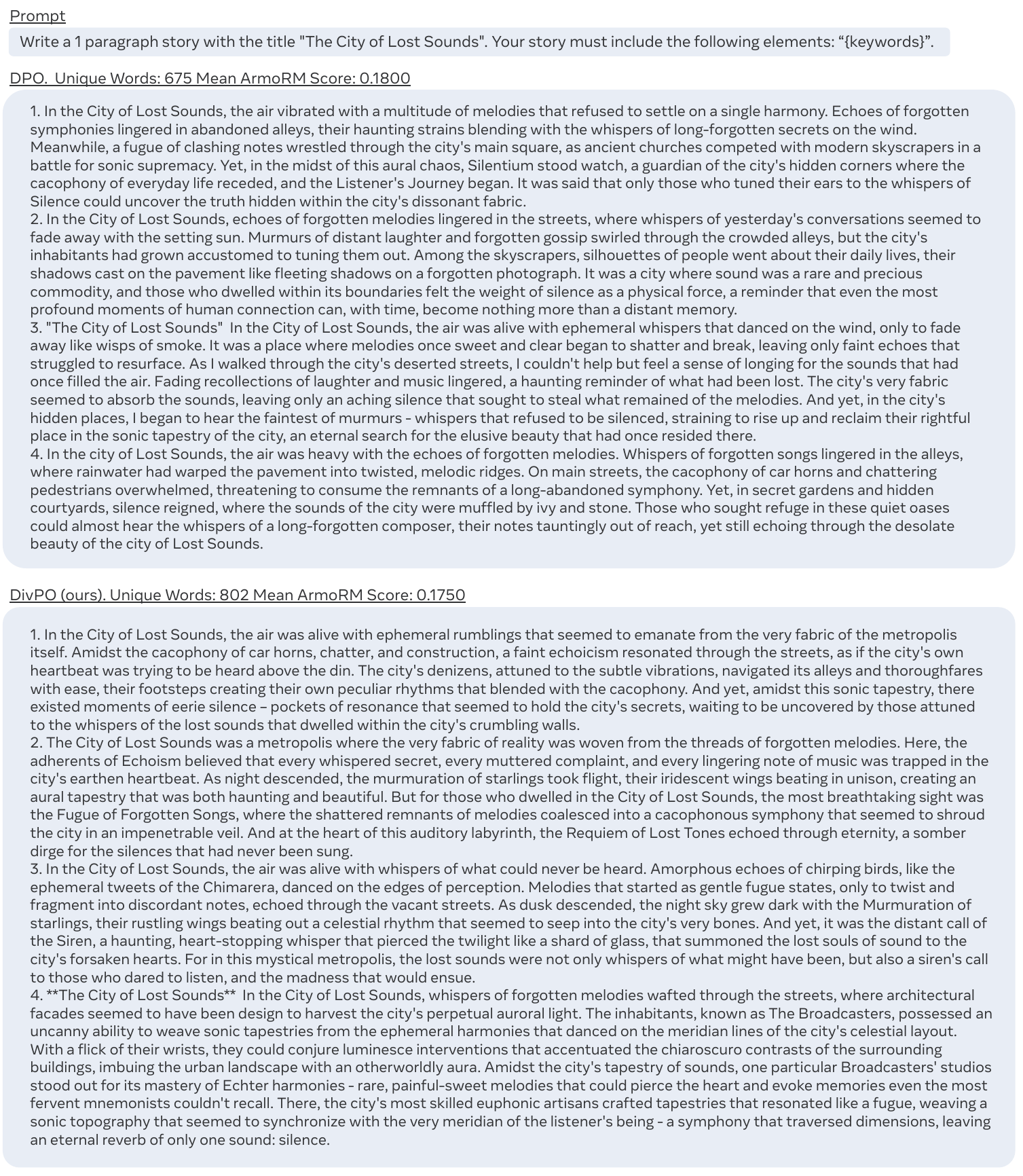}
    \caption{{\bf Full Story Example}. We show the first 4 stories generated by the DPO and \ourmethodsmall{} ($\mathcal{D}$=Prob, $\rho$=0.3) models. We report the Unique Words and Mean ArmoRM reward for all $N=16$ stories.}
    \label{fig:seededstory_full_example}
\end{figure*}

\section{Additional Persona Generation Task Results}

\begin{figure*}
    \centering
\includegraphics[width=0.9\linewidth]{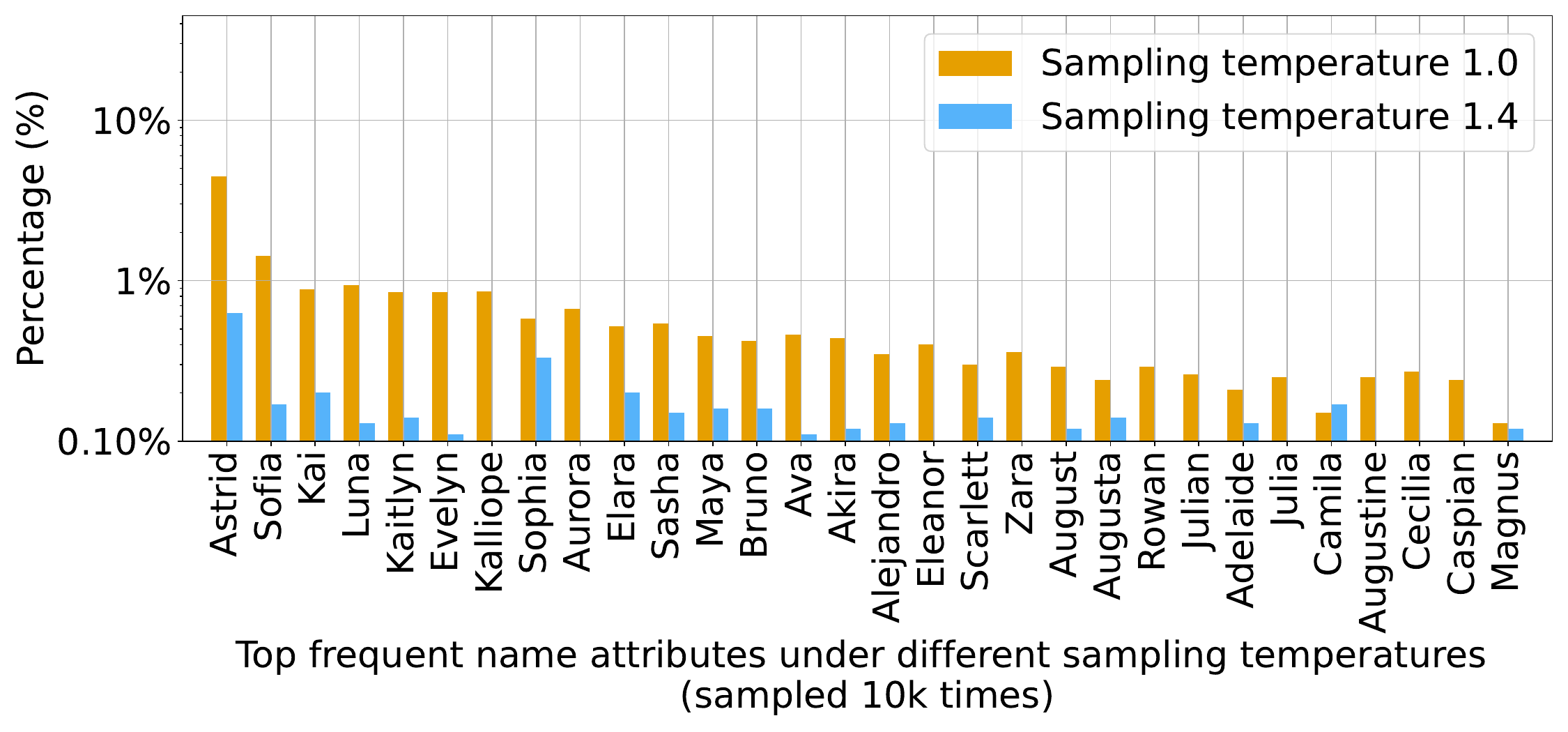}
    \caption{Counts (as percentage) of generated name attributes (computed only over valid JSON outputs) of persona for two temperature settings of Llama 3.1-8B-Instruct on the Persona Generation Task.}
\label{fig:persona_pretrained_llms_b}
\end{figure*}

~\autoref{fig:persona_pretrained_llms_b} shows the distribution of generated name attributes of personas for two different temperature settings (1.0 and 1.4) of Llama 3.1-8B-Instruct on the Persona Generation Task. We see that the higher temperature generates personas more uniformly.

\begin{figure*}
    \centering
    \includegraphics[width=0.8\linewidth]{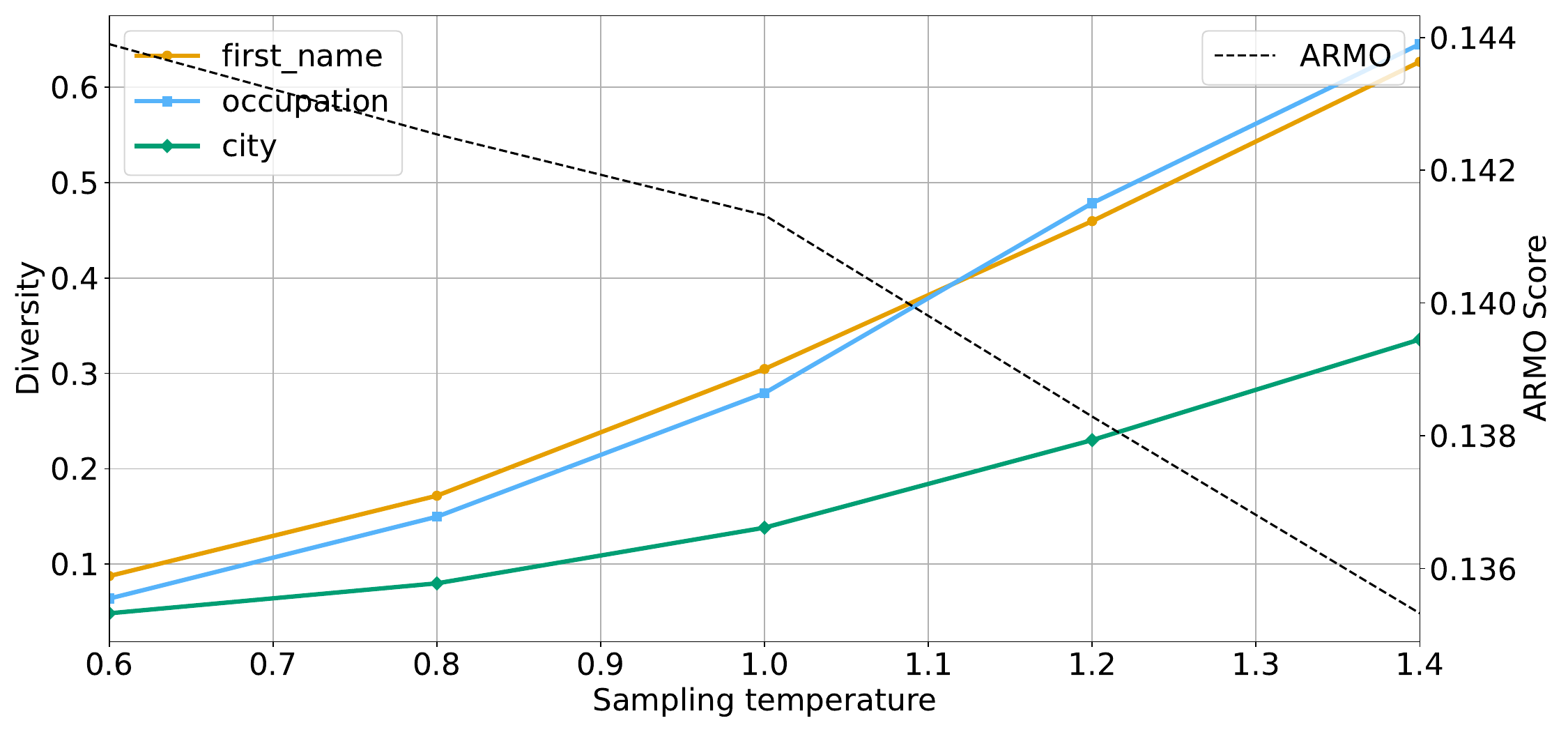}
    \caption{Diversity vs. quality trade-off of the Llama 3.1-8B-Instruct model as we change the sampling temperature in the Persona Generation Task.}
    \label{fig:persona_pretrained_llms_a}
\end{figure*}

However, in ~\autoref{fig:persona_pretrained_llms_a} we show the diversity vs. quality trade-off as we change the sampling temperature in the Persona Generation Task. We demonstrate that as we increase temperature, diversity increases, but quality (measured by ArmoRM score) decreases.

\begin{figure*}[ht]
    \centering
    \includegraphics[width=1\linewidth]{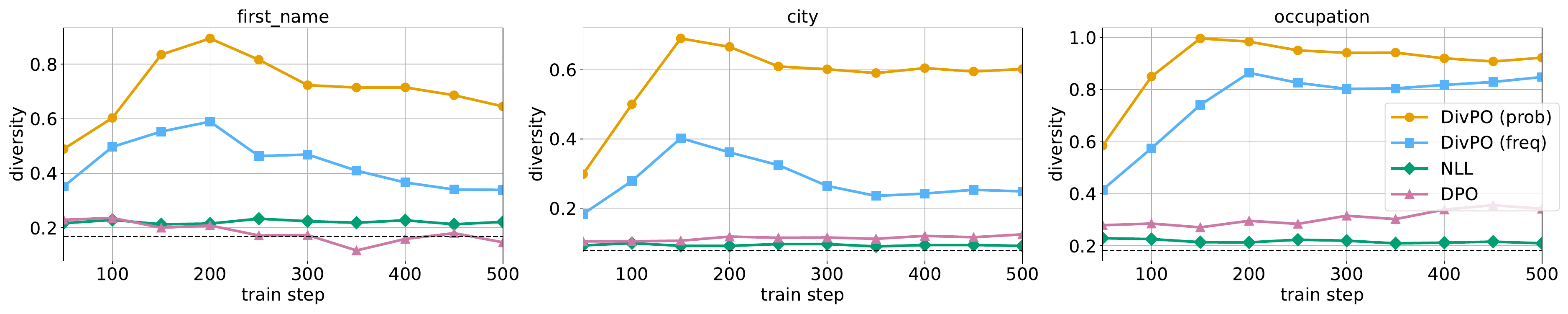}
    \includegraphics[width=1\linewidth]{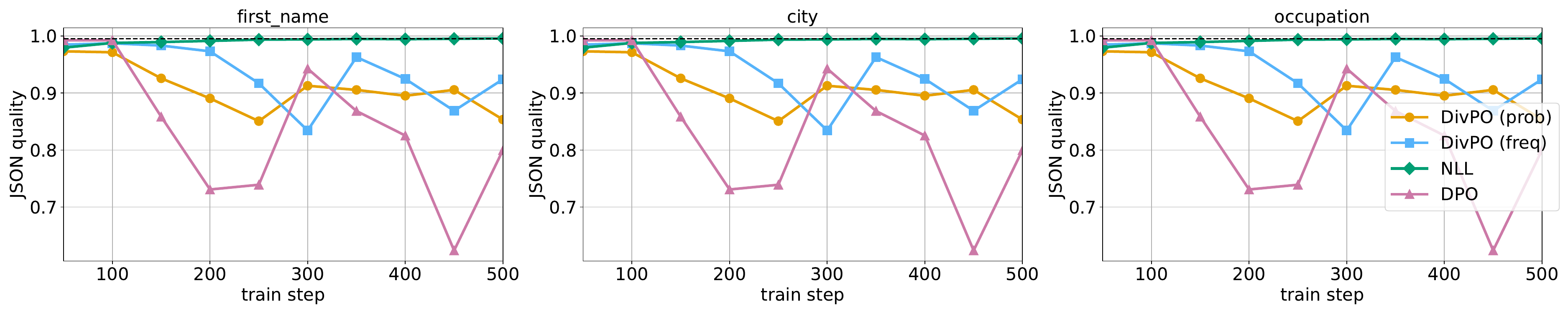}
    \includegraphics[width=1\linewidth]{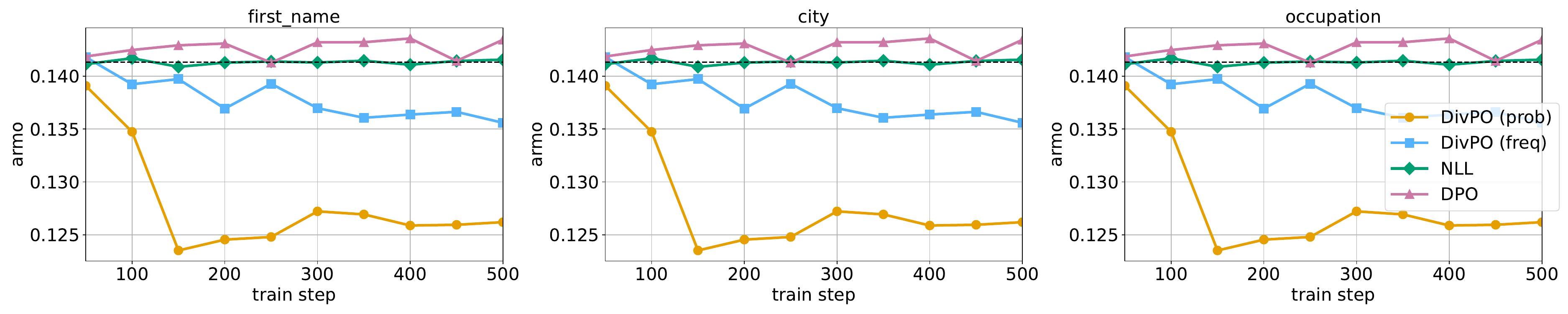}
    \includegraphics[width=1\linewidth]{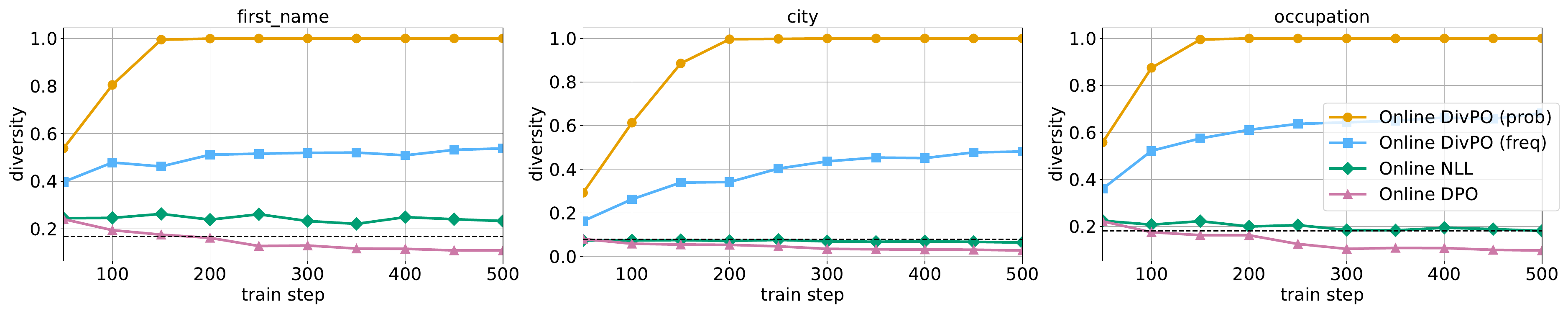}
    \includegraphics[width=1\linewidth]{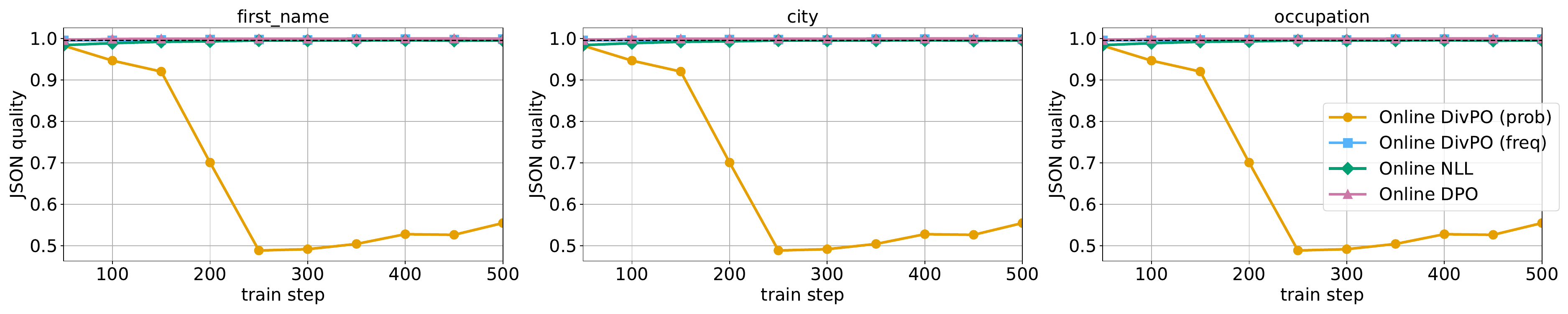}
    \includegraphics[width=1\linewidth]{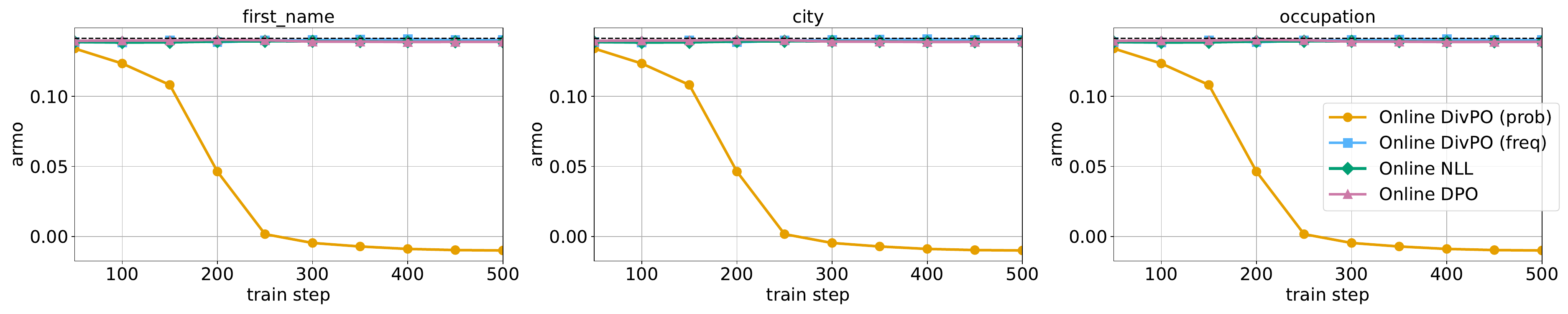}
    \caption{Diversity and quality evaluation in structured persona generation task. Dashed line shows performance of pretrained Llama 3.1 8B Instruct. All generations done with sampling temperature 1.0.}
    \label{fig:persona-divpo-results}
\end{figure*}

\begin{table*}[ht]
    \centering
    \caption{Model selection (see \autoref{subsec:personatask} for details) steps for each method in the Persona Generation Task.}
    \begin{tabular}{lc}
        \toprule
        Method & Model selection step \\
        \midrule
        SFT    & 500   \\
        DPO    & 50   \\
        DivPO (freq)    & 100   \\
        DivPO (prob)    & 50   \\
        Online SFT    & 400   \\
        Online DPO    & 400   \\
        Online DivPO (freq)    & 350   \\
        Online DivPO (prob)    & 50   \\
        \bottomrule
        
    \end{tabular}
    
    \label{tab:model_selection_step_persona}
\end{table*}

\begin{table*}[ht]
    \centering
    \caption{Hyperparameter values used during training.}
    \setlength{\tabcolsep}{1.75pt}
    \begin{tabular}{lcccccccccc}
        \toprule
        & \multicolumn{3}{c}{\textbf{Personas}} & \multicolumn{3}{c}{\textbf{Keyword Stories}} & \multicolumn{3}{c}{\textbf{Instruction Following}} \\

\cmidrule(lr){2-5} \cmidrule(lr){5-7} \cmidrule(lr){7-9}
\textbf{Method} &  \textbf{Batch Size} &  \textbf{Learning Rate} &  \textbf{} &  \textbf{Batch Size} &  \textbf{Learning Rate} &  \textbf{}  & \textbf{Batch Size} &  \textbf{Learning Rate} &  \textbf{}  \\
\midrule
        SFT & 16 & 1e-7 & & 64 & 1e-6 & & 64 & 1e-6  \\
        DPO & 16 & 1e-7 & & 64 & 1e-7 & & 64 & 1e-7 \\
        DivPO & 16 & 1e-7 & & 64 & 1e-7  & & 64 & 1e-7  \\
        \bottomrule
    \end{tabular}
    
    \label{tab:training_details}
\end{table*}

\section{Additional Instruction Following Experiments}
In ~\autoref{tab:alpacaeval_ablations} we ablate the number of generations $N$ for the instruction following task. We find that \ourmethodsmall{} is robust to varying $N$ values, and outperforms DPO for all values.

\begin{table}[htbp]
\centering
\caption{\textbf{Instruction Following Ablations.} Here, we ablate the number of samples per prompt, $N$ during training. We find that \ourmethodsmall{} is robust to changes in $N$, across all $\rho$ values.}
\label{tab:alpacaeval_ablations}
\resizebox{1.0\linewidth}{!}{%
\begin{tabular}{lccccccc}
\toprule
Method & {N} & {Compr Ratio $\uparrow$} & {Unique 1-Grams $\uparrow$} & {Entropy $\uparrow$} & {ArmoRM$\uparrow$} & {Len. (words)} \\
\midrule
DPO & 16 & 0.1949 & 1115.9 & 165.9 & 0.1803 & 283.4 \\
DivPO, $\mathcal{D}$=Prob, $\rho$=$0.1$ & 16 & 0.2107 & 1248.4 & 216.7 & 0.1814 & 288.4 \\
DivPO, $\mathcal{D}$=Prob, $\rho$=$0.2$ & 16 & 0.2187 & 1324.9 & 222.8 & 0.1804 & 285.9 \\
DivPO, $\mathcal{D}$=Prob, $\rho$=$0.3$ & 16 & 0.2370 & 1460.3 & 274.5 & 0.1794 & 295.0 \\
DivPO, $\mathcal{D}$=Prob, $\rho$=$0.4$ & 16 & 0.2501 & 1608.2 & 316.2 & 0.1764 & 298.2 \\
DivPO, $\mathcal{D}$=Prob, $\rho$=$0.5$ & 16 & 0.2632 & 1787.7 & 396.9 & 0.1699 & 301.9 \\
\midrule
DPO & 32 & 0.1873 & 1110.2 & 146.2 & 0.1792 & 283.9 \\
DivPO, $\mathcal{D}$=Prob, $\rho$=$0.1$ & 32 & 0.2206 & 1321.0 & 237.7 & 0.1801 & 292.0 \\
DivPO, $\mathcal{D}$=Prob, $\rho$=$0.2$ & 32 & 0.2232 & 1329.3 & 227.2 & 0.1794 & 280.6 \\
DivPO, $\mathcal{D}$=Prob, $\rho$=$0.3$ & 32 & 0.2427 & 1478.6 & 282.8 & 0.1780 & 288.7 \\
DivPO, $\mathcal{D}$=Prob, $\rho$=$0.4$ & 32 & 0.2519 & 1584.5 & 317.6 & 0.1776 & 295.9 \\
DivPO, $\mathcal{D}$=Prob, $\rho$=$0.5$ & 32 & 0.2851 & 2003.1 & 537.9 & 0.1620 & 304.2 \\
\midrule
DPO & 64 & 0.1936 & 1121.2 & 163.4 & 0.1791 & 286.3 \\
DivPO, $\mathcal{D}$=Prob, $\rho$=$0.1$ & 64 & 0.2205 & 1325.6 & 234.2 & 0.1798 & 288.6 \\
DivPO, $\mathcal{D}$=Prob, $\rho$=$0.2$ & 64 & 0.2395 & 1480.9 & 289.5 & 0.1790 & 296.7 \\
DivPO, $\mathcal{D}$=Prob, $\rho$=$0.3$ & 64 & 0.2625 & 1760.4 & 414.8 & 0.1698 & 307.3 \\
DivPO, $\mathcal{D}$=Prob, $\rho$=$0.4$ & 64 & 0.2706 & 1965.8 & 511.6 & 0.1442 & 301.6 \\
DivPO, $\mathcal{D}$=Prob, $\rho$=$0.5$ & 64 & 0.2965 & 2500.7 & 741.5 & 0.1244 & 316.3 \\
\bottomrule
\end{tabular}
}
\end{table}

\end{document}